%% file: main.tex
\documentclass[lettersize,journal]{IEEEtran}
\usepackage{amsmath,amsfonts}
\usepackage{algorithmic}
\usepackage{array}
\usepackage[caption=false,font=normalsize,labelfont=sf,textfont=sf]{subfig}
\usepackage{textcomp}
\usepackage{stfloats}
\usepackage{url}
\usepackage{verbatim}
\usepackage{graphicx}
\usepackage{cite}
\usepackage{caption}
\usepackage{comment}
\usepackage{multicol}
\newcommand{\kc}[1]{\textcolor{black}{#1}}
\newcommand{\pk}[1]{\textcolor{black}{#1}}
\newcommand{\ck}[1]{\textcolor{black}{#1}}

\input{preamble}

\def\rstgcn{\textsc{RSTGCN}\xspace}

\begin{document}
\title{RSTGCN: Railway-centric Spatio-Temporal Graph Convolutional Network for Train Delay Prediction}

\author{Koyena~Chowdhury,
        Paramita~Koley,
        Abhijnan~Chakraborty,
        Saptarshi~Ghosh
        
        \thanks{K. Chowdhury, A. Chakraborty, and S. Ghosh are with the Department of Computer Science and Engineering, Indian Institute of Technology Kharagpur, West Bengal - 721302, India. 
        P. Koley is with the International Institute of Information Technology, Bhubaneswar, Odisha - 752054, India and was partially supported by the SERB-National Postdoctoral Fellowship during this work.
        E-mail: koyenachowdhury02@gmail.com;
        paramita2000@gmail.com;
        abhijnan@cse.iitkgp.ac.in; saptarshi@cse.iitkgp.ac.in.}
  
}

\maketitle
\input{000abstract}


\input{010intro}

\input{011Related_Work}
\input{020Problem}

\input{040Experiment}
\input{050Conclusion}




%

\bibliographystyle{IEEEtran}

\input{main.bbl}
\newpage

%
%
%
%

\vfill

\end{document}

%% file: preamble.tex
\usepackage{booktabs}       
\usepackage{nicefrac}       
\usepackage{microtype}      
\usepackage{cleveref}
\usepackage{color}
\graphicspath{ {./Results/} }
\usepackage{multirow}
\usepackage{placeins}
\usepackage{float}
\usepackage[linesnumbered,ruled]{algorithm2e}
\usepackage{wrapfig}
\usepackage{lipsum}

\newcommand{\todo}[1]{\textcolor{red}{TODO: #1}}

\crefname{section}{Section}{Sections}

\newcommand{\our}{RSTGCN\xspace} 
\newcommand{\ourad}{RSTGCN$_{\text{AvgDelay}}$\xspace} 
\newcommand{\ouradh}{RSTGCN$_{\text{AvgDelay+Headway}}$\xspace} 
\newcommand{\ouradtd}{RSTGCN$_{\text{AvgDelay+TotDelay}}$\xspace} 
\newcommand{\ourbase}{RSTGCN$_{\text{Base}}$\xspace} 
\newcommand{\ourrelu}{RSTGCN$_{\text{ReLU}}$\xspace} 
\newcommand{\oursattn}{RSTGCN$_{\text{sattn}}$\xspace} 


%% file: 000abstract.tex
\begin{abstract}

Accurate prediction of train delays is critical for efficient railway operations.
While earlier approaches have largely focused on forecasting the exact delays of individual trains, studies on  {station-level delay prediction} is somewhat sparse. 
To address this gap, we propose the Railway-centric
Spatio-Temporal Graph Convolutional Network (RSTGCN), designed to forecast average arrival delays of all the incoming trains at a particular station for a particular time period. Our approach incorporates several architectural innovations and novel feature integrations, including train frequency–aware spatial attention, which significantly enhance predictive performance. 
To support this effort, we curate and release a comprehensive dataset for the entire Indian Railway Network (IRN), spanning 4,735 stations across 17 zones — the largest and most diverse railway network studied to date. 
We conduct extensive experiments using multiple state-of-the-art baselines, demonstrating consistent improvements across standard metrics. 
Specifically, RSTGCN outperforms the best baseline by 18\% in MAE, 14\% in MAPE, and 1–8\% in RMSE on the IRN.

\end{abstract}

%% file: 010intro.tex
\section{Introduction}
\label{sec:intro}

\noindent Indian Railways is one of the largest railway systems globally, spanning approximately 69,000 kilometers and serving as the backbone of long-distance transportation in India. 
It facilitates the daily movement of more than 23 million passengers across India. 
Indian Railways plays a pivotal socio-economic role, enabling access to employment, education, healthcare, and markets for a significant portion of the population.

However, the vast scale and operational complexity of this mass transit system contribute to one of its most persistent challenges, \textit{i.e.,} \textit{unpredictable train delays}. Although following schedules is essential for maintaining the reliability of the railway network, various operational challenges, such as network congestion, adverse weather, infrastructure constraints, and the coexistence of multiple types of trains sharing the same tracks, unlike in most other countries, often lead to significant delays.
\kc{Also, several external factors, such as disruptions caused by large passenger movements during festivals, examinations and office-hours, chain pulling incidents, social or political disruptions, trespassing, and accidents may cause huge delays. In addition, the frequent introduction of new trains, often scheduled based on increasing passenger demand rather than timetable optimization, can further exacerbate congestion and cumulative delay. In India, generally punctuality yardstick is considered as 15 minutes for express trains.}
However, our analysis of Indian Railways data for the month of September 2024 underscores that the delays are often much more severe, revealing an average arrival delay of \ck{ approximately 32 minutes, and several instances extending beyond 12 hours. Notably, 25\% of delays exceed 30 minutes, 13\% exceed 1 hour, and 7\% exceed 2.5 hours.} 

\IEEEpubidadjcol

Disruptions in train timing are often interconnected: a single out-of-schedule train can trigger a ripple effect, delaying other trains and congesting adjacent stations. 
In such a heavily utilized, high-traffic, and interdependent system, accurate delay prediction can enable real-time risk assessments, dynamically adjust schedules, and mitigate delay propagation. 
In addition, it offers meaningful insight into the operational behavior of the rail network, providing a foundation for long-term strategic planning and improvements.

\kc{While multiple prior works have looked at predicting train-specific delays~\cite{li2024bayesian, li2024railway, huang2020deep, huang2020modeling}, \textit{forecasting station-specific aggregated delay} can also provide great operational value, particularly in supporting station-centric tasks and policies, such as train dispatch and control decisions~\cite{zhang2021train, zhang2022interpretable, heglund2020railway}. 
Hence, in this work, 
we address the challenge of \textit{forecasting the average arrival delay at each station over a short-term future horizon}.
} 
Notably, prior station-based delay prediction studies have primarily been conducted in the context of the high-speed Chinese railway network~\cite{zhang2021train, zhang2022interpretable}, where generally higher levels of punctuality lead to a modeling emphasis on the count of delayed trains per hour. In contrast, the Indian Railway context presents a different operational dynamics, where predicting the magnitude of delays is more relevant than simply identifying delayed instances. Accordingly, we shift the focus to modeling average arrival delays per hour. A similar formulation has also been explored in the context of the British railway network~\cite{heglund2020railway}.

\vspace{2mm}
To address the task of predicting arrival delays, we make several important contributions in this paper: \\
(1) We propose a Graph Convolutional Network (GCN) based spatio-temporal framework for station-wise hourly delay prediction, 
which we name
\textit{\textbf{R}ailway-centric \textbf{S}patio-Temporal \textbf{G}raph \textbf{C}onvolutional \textbf{N}etwork} (\rstgcn).
We incorporate several domain-informed features and key architectural enhancements in our proposed model.
In particular, we introduce hourly headway, a novel traffic-based feature, along with average and total arrival/departure delays. 
We also redesign the spatial attention module to integrate the train frequencies between stations and adjust the output activation for modeling delay values rather than delay counts. 
When evaluated on real-time large-scale temporal data from Indian Railways, our model improves over several state-of-the-art baselines.\\
(2) We present the first nation-wide train delay dataset for Indian Railways comprising $3,892$ long-distance trains and $4,735$ stations. This dataset is orders of magnitude larger than most existing railway traffic datasets; e.g., compared to 727 stations in~\cite{zhang2021train} and 9 stations in~\cite{li2024bayesian}, our dataset has  $4,735$ stations.
Our dataset is available for research purposes at \url{https://github.com/KoyenaChowdhury/RSTGCN/}.
\\
%
(3) We would also like to highlight that while prior works on delay prediction in the Indian Rail network have focused on only a single route~\cite{pradhan2021simulating} or a partial region~\cite{gaurav2018estimating}, to the best of our knowledge, this is the first work to study delay prediction over the entire IRN, as well as over individual zones.

%% file: 011Related_Work.tex
\section{Related Work}
\label{sec:related_work}

Existing literature on train delay prediction can be broadly classified into three categories, namely mathematical model-driven methods, statistical, and machine learning methods~\cite{spanninger2022review}.

\ck{\textit{Mathematical model-driven methods} primarily adopt stochastic and analytical formulations to characterize delay propagation. Representative works use probabilistic distributions to estimate knock-on effects \cite{meester2007stochastic, yuan2007optimizing}, and examine signaling-constrained recovery and capacity limits \cite{goverde2013railway, ranjbar2022impact}. Further advancements include max-plus algebra \cite{goverde2010delay}, adaptive event graphs \cite{kecman2014online}, and epidemic dynamics \cite{qin2025study} to model network-wide propagation. 
These methods are not directly comparable to our work, since their objectives are different. For instance,  \cite{meester2007stochastic} focuses on delay distributions and phase-type propagation analysis, \cite{yuan2007optimizing} focuses on station-capacity optimization under signaling constraints, \cite{goverde2013railway, goverde2010delay} on signaling and capacity-oriented analyses, \cite{ranjbar2022impact, kecman2014online} on event-graph-based conflict and propagation modeling, and \cite{qin2025study} on statistical characterization of delay propagation rather than delay prediction.}

\textit{Statistical methods} are primarily developed around Bayesian network based framework, exploring 
the effect of prediction horizon and train information~\cite{corman2018stochastic}, complexity and dependency of train operations~\cite{lessan2019hybrid}, modeling delay propagation and clustering delay patterns~\cite{huang2022enhancing}, along with other approaches as fuzzy petri-nets~\cite{milinkovic2013fuzzy}, which is based on the Markov assumption. \pk{Among recent Markov chain based works, Shahin et al.~\cite{csahin2017markov} explored train delay prediction with homogeneous Markov chains, while Xu et al.~\cite{xu2022novelmarkovmodelnearterm} proposed a delay prediction model based on non-homogeneous Markov chain with  sparse transition matrix modeled by Gaussian kernel density estimation. Spanninger et al.~\cite{spanninger2023train}, and Artan et al. \cite{artan2025comparative} further employed advanced Markov chains setting based on process time deviations instead of absolute delays.
}

Finally, \textit{machine learning models} employed various modeling paradigms for addressing train delay prediction, including decision tree~\cite{lee2016delay}, ANN~\cite{yaghini2013railway}, shallow and deep extreme learning machines~\cite{oneto2018train}. 
Among the sequential models, Wen et al.\cite{wen2020predictive} used LSTM to model delay-influencing factors on Dutch railways. Temporal-only approaches, such as Ping et al.\cite{ping2019neural}, leverage RNNs, while Oneto et al.\cite{oneto2016advanced} employ data-driven models based solely on historical data, overlooking spatial correlations.
Huang et al.~\cite{huang2020modeling, huang2020deep} proposed FCLL-Net and CLF-Net, combining LSTM, CNN, and FNN to capture temporal, spatio-temporal, and external operational features. Recently, Li et al.~\cite{li2023prediction} proposed TLF-net to capture the effects of rolling stock connections and terminal route conflicts.
Recently, Zhang et al. \cite{zhang2024multi} employed a temporal point process–based framework (TANTPP); Gao et al. \cite{gao2024data} 
applied XGBoost algorithm.

While the above works ignore the spatial dependency in the network, recent works employ tools from graph neural network literature to exploit complex spatio-temporal dependency.
Among them,  
GCRNN~\cite{seo2018structured} and  STGCN~\cite{yan2018spatial} combine GCNs with recurrent or temporal convolution layers. ASTGCN~\cite{guo2019attention} introduced attention mechanisms, while TSTGCN~\cite{zhang2021train} further incorporated inter-station distances. 
Heglund et al.\cite{heglund2020railway} applied STGCN to the British railway; whereas Li et al.\cite{li2024railway} introduced SAGE-Het for modeling heterogeneous interactions. 
\kc{Zhang et al. proposed MATGCN~\cite{zhang2022prediction}, followed by C-MATGCN~\cite{zhang2022interpretable} using graph communities and fuzzy trees. 
MATGCN~\cite{zhang2022prediction} combined some external factors, such as weather and holidays, using DLP and employed a multifeature attention mechanism to capture their impact on train delays.} 
DB-STGCN\cite{li2024bayesian} incorporated domain knowledge via Bayesian causality graphs. 
\pk{
Among these prior works, \cite{heglund2020railway, zhang2021train, zhang2022prediction, zhang2022interpretable} focus on station-specific delay prediction, while the others are train-specific. 
Note that, we exclude~\cite{zhang2022prediction, zhang2022interpretable, li2024bayesian} 
from baseline comparisons due to their reliance on weather data, which is not available for IRN.
}

Traffic flow prediction is similar well-studied problem in spatio-temporal modeling, whose solutions often apply to train delay prediction without major modifications. 
Early approaches include spatio-temporal GCNs~\cite{yu2018spatio} and attention mechanisms for long-term dependencies~\cite{guo2019attention}.
Further developments have explored dynamic spatial modeling through Laplacian estimators~\cite{diao2019dynamic}, temporal GCNs~\cite{zhao2019t}, adaptive graph learning~\cite{zhang2022adapgl}, and the integration of networking principles~\cite{farreras2023improving}.
Recent advances include STMGCN~\cite{liang2022fine} for multi-correlation modeling, STCL-AGA~\cite{zhang2024spatio} with graph augmentation and masking, DAGCRN~\cite{shi2023dagcrn} with adaptive adjacency and temporal attention, STGSA~\cite{wei2023stgsa} for localized and long-term dependency capture, and STFTGCN~\cite{zhang2025trend}, which integrates trend-aware spatio-temporal features.

%
Most of the above train delay prediction works focus on the Chinese rail network, and there has been few attempts on analyzing IRN. Among them, the notable works include statistical delay analysis for express trains~\cite{ghosh2013run}, arrival delay prediction via a zero-shot framework in partial region with 135 trains~\cite{gaurav2018estimating}, and delay prediction in the Delhi-Mumbai route~\cite{pradhan2021simulating}. However, none of these works addressed delay prediction across the entire Indian Railway network or different zones, highlighting the need for a system-wide study.

Note that, none of the above approaches takes into account the influence of train counts between stations on delay propagation. To bridge this gap, we enhance the spatial attention of the GNN and also introduce hourly headway as a novel feature in the present work. 

%% file: 020Problem.tex
\section{Dataset description}
\label{sec:data}


We develop a dataset based on train operation records from the Indian Railway (IR) system, including data from $3,892$ long-distance passenger trains (excluding freight trains and `local' trains which travel only in the vicinity of metropolitan cities). 
This dataset is sourced from \url{https://runningstatus.in/}, which provides detailed information such as running date, train number, train name, station code, station name, distance from source, scheduled and actual arrival and departure times along with delays for various express trains over a specified time period, as shown in Table~\ref{tab:train_records}.
The train running information spans the period September 01-30, 2024. \kc{We chose September as a representative month as it avoids extreme climate conditions (neither peak-monsoon, nor the winter period), and is without any major festivals in India. 
Our dataset contains 1,282,326 train records, where each record corresponds to a train at a specific station on a given day. We exclude instances where no train is scheduled or where no delay is recorded for the corresponding time interval.}\\

\begin{table*}[tb]
\centering
\caption{Examples of train running records in our dataset}
\resizebox{\textwidth}{!}{
\begin{tabular}{l l l l l l l l l l l l l}
\hline
\textbf{Date} & \textbf{Train No.} & \textbf{Train Name} & \textbf{Code} & \textbf{Station} & \textbf{Dist.} & \textbf{Sch. Arr.} & \textbf{Act. Arr.} & \textbf{Arr. Delay} & \textbf{Sch. Dep.} & \textbf{Act. Dep.} & \textbf{Dep. Delay} \\
\hline
7 Sept. 2024 & 12277 & Satabdi Express & KGP & Kharagpur & 116 KM & 03:55 PM & 04:19 PM & 24M Late & 04:00 PM & 04:24 PM & 24M Late \\
7 Sept. 2024 & 12277 & Satabdi Express & BLS & Balasore  & 237 KM & 05:15 PM & 05:35 PM & 20M Late & 05:20 PM & 05:40 PM & 20M Late \\ [2pt]
25 Sept. 2024 & 12951 & NdlsTejasRaj Exp. & BVI & Borivali  & 30 KM & 05:22 PM & 05:40 PM & 18M Late & 05:24 PM & 05:42 PM & 18M Late \\
25 Sept. 2024 & 12951 & NdlsTejasRaj Exp. & ST & Surat  & 262 KM & 07:43 PM & 07:53 PM & 10M Late & 07:48 PM & 07:58 PM & 10M Late \\
\hline
\end{tabular}
}
\label{tab:train_records}
\end{table*}

\noindent \kc{\textbf{The Indian Railways Network dataset:}
We represent the IR system by a network (which we call the IR network or IRN) where the nodes/vertices are the stations, and there is an edge between two nodes $s_i$ and $s_j$ if and only if they are \textit{consecutive} stations on at least one train route. In other words, there is a link/edge $(s_i, s_j)$ only if there is a direct track/line connecting the two stations.}
The IRN thus formed comprises $4,735$ stations (the major train stations in India) connected by $9,336$ edges (direct lines). 
The key statistics of the IR network in our dataset are summarised in Table~\ref{tab:raildata}.

\begin{table}[h]
\centering
\scriptsize
\caption{Statistics of Indian Railways}
\begin{tabular}{l c | l c}
\hline
\textbf{Description} & \textbf{Value} & \textbf{Description} & \textbf{Value} \\
\hline
No. of Stations & 4735 & No. of Trains & 3892 \\
No. of Edges & 9336 & Avg Node degree & 3.94 \\
Avg Trains per Station & 17.95 & Avg Station Halts per Train & 21.84 \\
Avg. distance (KM) & 38.05 & Avg. trains per edge & 8.69 \\
Avg Headway (mins) & 48.76 & Median Headway (mins) & 14.0 \\
\hline
\end{tabular}
\label{tab:raildata}
\end{table}

\begin{figure}[h]
\centering
\begin{minipage}{0.40\linewidth}
    \centering
    \includegraphics[width=\linewidth]{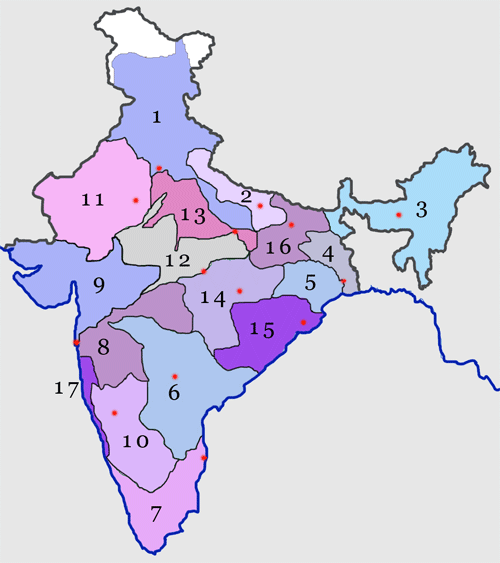}
    \caption{\kc{Zones in the\\ Indian Railways}}
    \label{fig:zones}
\end{minipage}
\begin{minipage}{0.50\linewidth}
    \centering
    \scriptsize
    \begin{tabular}{l}
    \textbf{\underline{Zone (Code) - Number of Stations}}\\
    1. Northern Railway (NR) - 449\\
    2. North Eastern Railway (NER) - 218\\
    3. Northeast Frontier Railway (NFR) - 286\\
    4. Eastern Railway (ER) - 236\\
    5. South Eastern Railway (SER) - 205\\
    6. South Central Railway (SCR-499) \\
    7. Southern Railway (SR) - 316\\
    8. Central Railway (CR-) 272\\
    9. Western Railway (WR) - 424\\
    10. South Western Railway (SWR) - 237\\
    11. North Western Railway (NWR) - 334\\
    12. West Central Railway (WCR) - 216\\
    13. North Central Railway (NCR) - 240\\
    14. South East Central Railway (SECR) - 113\\
    15. East Coast Railway (ECOR) - 264\\
    16. East Central Railway (ECR) - 354\\
    17. Konkan Railway (KRCL) - 72\\
    \end{tabular}
\end{minipage}
\end{figure}

\noindent \textit{Zones in the IRN:}
For organizational convenience at different levels, the IR network is divided into $17$ major regions called `zones'~\cite{jose2023multi}. 
This zonal division allows the extremely large system to balance technical, geographic, financial, and organizational factors, and make the system responsive to regional needs.
Fig~\ref{fig:zones} gives a map of the IRN showing the zones, and  states the name and number of stations of each zone. 
The zones vary in size, ranging from South Central Railway (SCR) with $499$ stations to Konkan Railway Corporation Limited 
with only $72$ stations.
Our model is evaluated both on the full network and separately for each zone, to ensure robust validation across diverse network scales.

\section{Problem statement}
\label{sec:problem}

A railway network is defined as a undirected graph $G = (S, E, A, M, K)$, where $S$ denotes the set of stations with $|S| = N$, $E$ denotes set of edges, where each edge denotes existence of a track between two consecutive stations. $A$ is the adjacency matrix which denotes connectivity between stations. 
$M$ denotes the distance matrix between stations. 
$K$ is the train frequency matrix, 
where $k_{ij}$ denotes the number of distinct trains between the adjacent stations $i$ and $j$.
%
Each train $r$ has a schedule that specifies exact arrival time $t_r^a$ and departure time $t_r^d$ for each station. 
Suppose, the train actually reaches the station at $\bar{t}_r^a$ and departs at $\bar{t}_r^d$. Then arrival delay $\Delta_a$ is defined as the difference between actual arrival time and scheduled arrival time, 
$\Delta_a = \bar{t}_r^a - t_r^a$. Similarly, departure delay is defined as the difference between actual departure time and scheduled departure time, $\Delta_d = \bar{t}_r^d - t_r^d$. 
In our work, we are particularly interested in \textit{hourly arrival delay ($\Delta_a$) prediction at each station averaged across the trains arriving in a particular time period (e.g., a particular hour)}.
\ck{Note that, if one desires to differentiate between minor deviations and operationally meaningful delays, appropriate thresholds on $\Delta_a$ and $\Delta_d$ can be considered. For instance, one can consider a train to be delayed only if $\Delta_a$ is greater than 5 or 10 minutes.}

Given a fixed time period \( \tau \), let \( F \) be the number of features per station (for our method, \( F=5 \)). Each station records several features capturing various train running statistics during \( \tau \), 
which we will discuss in detail in the following section \ref{sec:features}. The feature matrix for all stations is denoted as \( X_\tau = (X_\tau^1, X_\tau^2, \ldots, X_\tau^N) \in \mathbb{R}^{N \times F} \), where \( X_\tau^i \in \mathbb{R}^F \) represents the feature vector of station \( i \). We set \( y_t^i \in \mathbb{R} \) as the average arrival delay at station \( i \) in the future time period \( t \). 

\vspace{2mm}
\noindent \textit{Problem:} 
Given a railway network $G$, the objective is to predict the average arrival delay at all stations in the network for the next \( t_p \) time steps. \( Y = (y_1, y_2, \ldots, y_N)^t \in \mathbb{R}^{N \times t_p} \), based on the historical and node-specific data of all stations $S$ over a past time period \( \tau \). Here, \( y_i = (y^{\tau+1}_i, y^{\tau+2}_i, \ldots, y^{\tau+t_p}_i) \) represents the predicted average delay sequence at station \( i \) over the future time window of length \( t_p \).

\section{\our: Proposed framework}

In this section, we propose \our  (Railway-centric Spatio-Temporal Graph Convolutional Network) 
for predicting the node-wise average delay in a rail network. 
Our model borrows some components from TSTGCN~\cite{zhang2021train}, but modifies the components/features to make them more suitable for the Indian Railway network.
We start by describing the station-specific features that we propose to incorporate in our model.

\subsection{Station-specific features}
\label{sec:features}
As our primary task is to predict station-specific average arrival delay, we design station-level features that directly capture delay magnitude rather than delay incidence. For each station, we maintain two primary features: (i) hourly average arrival delay and (ii) hourly average departure delay. This differs from prior work, which focused on counting delayed trains per hour~\cite{zhang2021train}. This formulation aligns with the regression nature of our task by modeling delay intensity instead of delay frequency. 
We further argue that not only average delays, but also \textit{total hourly arrival and departure delays} at each station play an important role in delay modeling. So, for each station, we also maintain total hourly arrival and departure delays.

Finally, we introduce a novel station-specific feature, called \textit{hourly headway}, defined as the mean scheduled time gap between consecutive trains at a station within a given hour. Prior studies have shown that train delays can be affected by headway~\cite{huang2024explainable}. We hypothesize that shorter headways constrain recovery time between successive trains, thereby amplifying delay persistence, whereas longer headways provide operational slack that facilitates delay absorption.

\subsection{Model details}
\begin{figure}[h]
     \centering {\includegraphics[width=.98\linewidth]
     {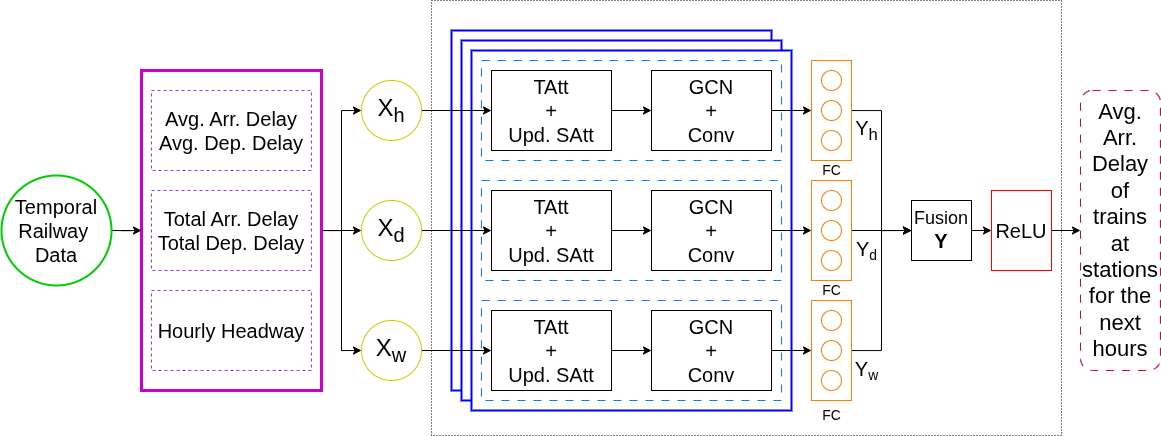}}
     \caption{RSTGCN model architechture}
     \label{fig:model}
 \end{figure}

Our framework comprises three distinct modules that capture recent, daily, and weekly temporal patterns, respectively. Each module contains multiple spatio-temporal encoder blocks, where every block sequentially integrates an attention layer, a convolution layer, and a fully connected layer, as illustrated in Fig.~\ref{fig:model}. The representations learned from the three temporal views are then linearly fused to generate the final prediction. In the following sections, we describe the architecture in detail, with particular emphasis on the key innovations of the proposed framework.

 \begin{figure*}[ht]
     \centering	
      \subfloat[]
      {\includegraphics[width=0.33\linewidth]
     {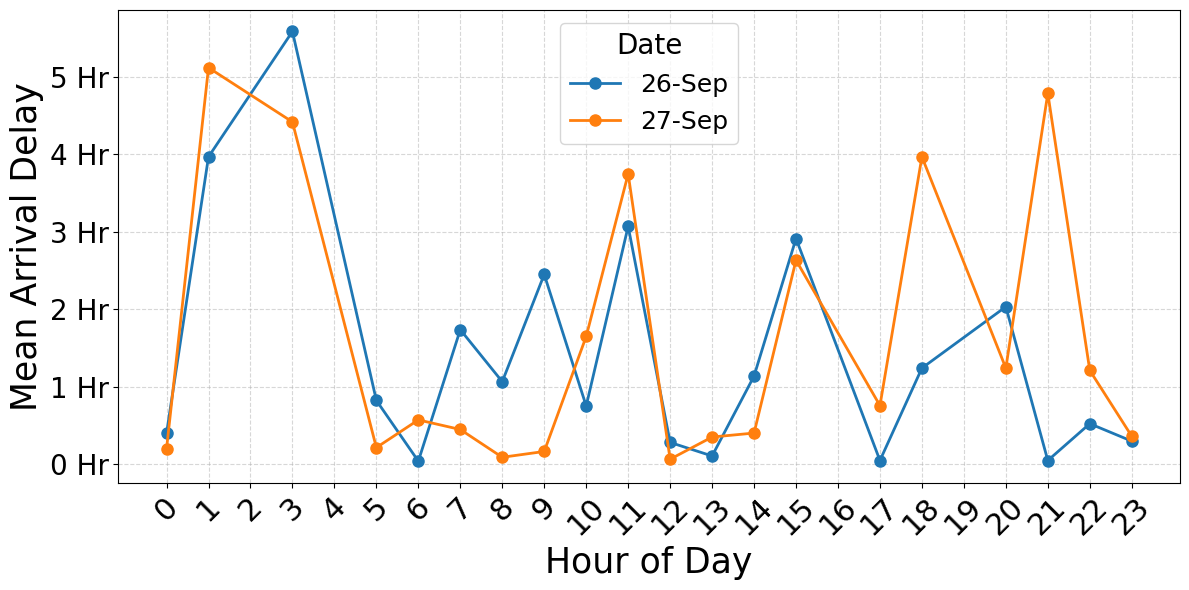} \label{fig:Aspatiotemp}}
      \subfloat[]      {\includegraphics[width=0.33\linewidth]
      {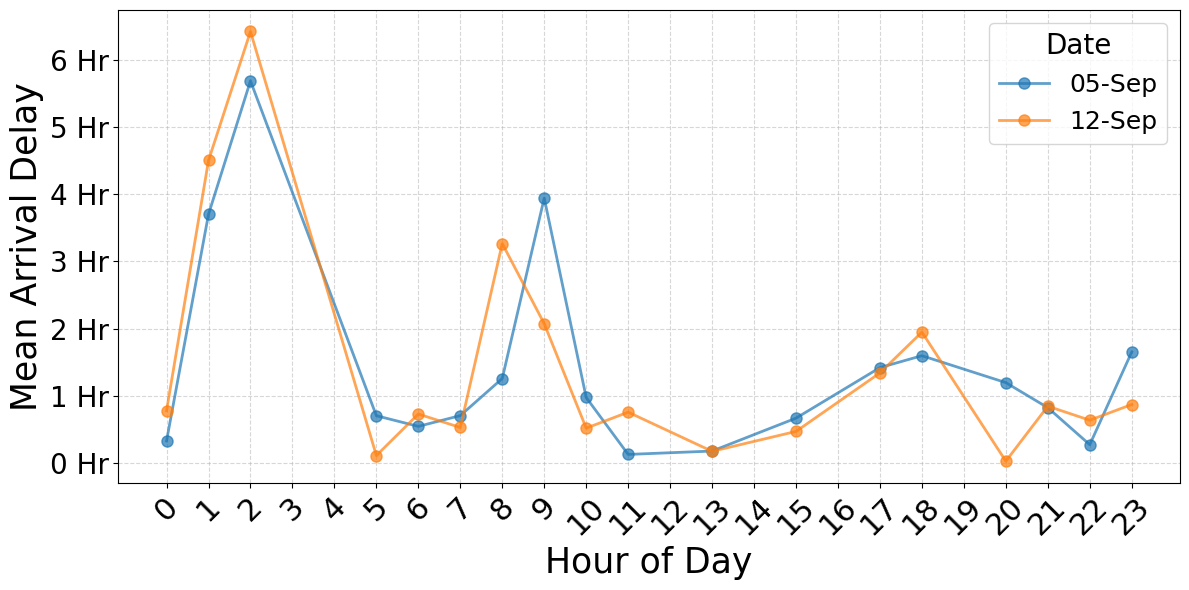} \label{fig:Bspatiotemp}}
      \subfloat[]      {\includegraphics[width=0.33\linewidth]
      {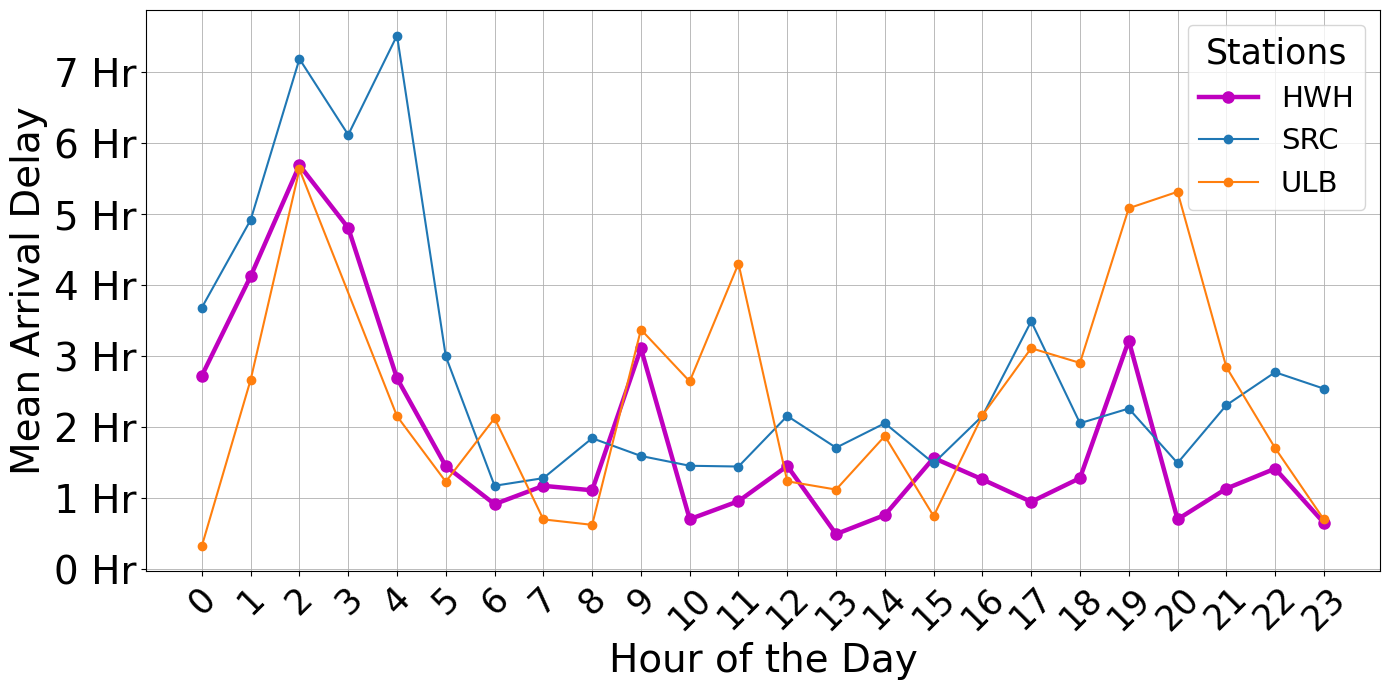} \label{fig:Cspatiotemp}}
     \caption{ \kc{\textbf{(a)} \textit{Daily (temporal) correlations:} Mean arrival delay (hourly) over two consecutive days at Howrah (HWH) Junction, 
     \textbf{(b)} \textit{Weekly (temporal) correlations:} Mean arrival delay (hourly) across two Thursdays in consecutive weeks at HWH Junction. 
     \textbf{(c)}~\textit{Spatial correlation:} Mean arrival delay (hourly) of HWH and some neighbouring stations. High delays at HWH often correspond to high delays at some adjacent station. While Figure 3a and 3b justify the usage of temporal features (daily and weekly history), Figure 3c advocates for the use of spatial attention.}}
     \label{fig:spatiotemp}
 \end{figure*}

\noindent \textit{Processing input data:}
The training data consists of the average arrival delay at the nodes of the network at different timestamps, which is a common time-series data in a graph network. With sampling frequency $q$ per day and prediction window $t_p$, we extract past graph signals $X_{\tau}$ over $t_h$, $t_d$, and $t_w$ as inputs to the recent, daily, and weekly components. The temporal timestamp inputs are described as follows: \\
\textit{(1) Recent history:} The arrival delay of a train at one station increases the chance of higher arrival delay of upcoming trains at that station. To encode this, for each node, we take as input the recent history, formed by the average arrival delay at previous $\tau$ hours at that node. From the network perspective, the recent history is expressed as:
\[
\mathbf{X}_h = (X_{t_0 - t_h + 1}, X_{t_0 - t_h + 2}, \ldots, X_{t_0}) \in \mathbb{R}^{N \times F \times t_h}
\]
\textit{(2) Daily history:} Train delays exhibit strong time-of-day dependence due to recurring traffic patterns in the railway network. Since many trains operate daily with fixed schedules, traffic conditions at a given hour tend to be similar across days, influencing arrival delays at that time. For example, Fig.~\ref{fig:Aspatiotemp} illustrates the mean arrival delays at Howrah (HWH) Junction, a major station in IRN, on two consecutive days. The similar trends observed across several time instances indicate strong temporal regularity in delay patterns, despite occasional variations. Therefore, the daily average delay at hour $t$ is incorporated as input:
\begin{multline*}
\mathbf{X}_d = (X_{t_0 - (t_d / t_p) \cdot q + 1}, \ldots, 
X_{t_0 - ((t_d / t_p) - 1) \cdot q + 1}, \ldots, \\
X_{t_0 - q + 1}, X_{t_0 - q + t_p}) 
\in \mathbb{R}^{N \times F \times t_d}
\end{multline*}
\textit{(3) Weekly history:} In India, many express trains operate on a weekly schedule, making weekly temporal patterns important for delay prediction. Fig.~\ref{fig:Bspatiotemp} illustrates the mean arrival delays at Howrah (HWH) Junction on September 5 and 12, 2024, where delay peaks occurring at specific hours on one day reappear at similar hours on the corresponding day of the following week, demonstrating strong weekly temporal dependence. 
Towards that, the following weekly history is taken into account to capture weekly rail traffic characteristics: 
\begin{multline*}
\mathbf{X}_w = \big( X_{t_0 - 7 \cdot (t_w / t_p) \cdot q + 1}, \ldots, 
X_{t_0 - 7 \cdot ((t_w / t_p) - 1) \cdot q + 1}, \ldots, \\
X_{t_0 - 7 \cdot q + t_p} \big) 
\in \mathbb{R}^{N \times F \times t_w}
\end{multline*}
For example, to predict the next 3 hours of delay (\(t_p = 3\)) at a particular station on Sunday, 21\textsuperscript{st} September at 8:00 AM, say, we set \(t_h = 3\), \(t_d = 1\), and \(t_w = 1\). This means the model will use the past 3 hours of delay data from 5:00 AM to 8:00 AM on the same day, the delay data from the same hour (8:00 AM) on the previous day, \textit{i.e.}, Saturday, 20\textsuperscript{th} Sept., and the delay data from 8:00 AM on the previous Sunday, \textit{i.e.}, 14\textsuperscript{th} September, all for the same station.

\vspace{2mm}
The input is then passed through the following layers for further processing.

\vspace{2mm}
\noindent \textit{Spatio-temporal attention:} 
This module captures long and short-term spatio-temporal dependencies. Generally, several layers of multi-head attention modules are employed here, as follows.

(1) \textit{Temporal attention:} 
This module computes the effects of historical delay at a specific node at various temporal granularities. 
We use a time-based self-attention mechanism, where the time weight matrix \( Z \) captures dependencies between time steps \( i \) and \( j \).
\[
Z = V_t \cdot \sigma\left( \left( (X U_1) U_2 \right) \odot (U_3 X) + b_t \right) 
\]
where \( X = (X_1, X_2, \ldots, X_{T_{r-1}}) \) is the input to the \( r \)-th spatio-temporal module. \( F_{r-1} \) and \( T_{r-1} \) denote the number of features and time steps at input layer \(r\), respectively. \( V_t \) and \( b_t \) are learnable projection parameters. Finally we get \( Z' \) by applying $Softmax$.
This temporal attention matrix \( Z' \) is next forwarded to the input of the \( r \)-th layer of the spatial attention module.

(2) \textit{Spatial attention:} 
The output of the temporal attention block is subsequently passed to the spatial attention module. Since delays at one station often propagate to neighboring stations over subsequent time steps, modeling spatial dependencies is essential. \kc{Fig.~\ref{fig:Cspatiotemp} illustrates this phenomenon for HWH Junction and nearby stations, where major delay spikes at HWH frequently coincide with similar spikes at adjacent stations, indicating strong spatial correlations.}
To capture these dependencies, we employ a spatial attention mechanism that jointly models inter-station proximity and train connectivity. A correlation matrix (C) is first computed as
\[
C = V_S \cdot \sigma\left( \left( (X_{Z'} \cdot W_1) W_2 \right) \odot (W_3 X_{Z'}) + b_S \right)
\]
where \( X_{Z'} \) denotes the input data from the time attention module, and \( W_1 \), \( W_2 \), and \( W_3 \) are feature transformation matrices.


The distance weight matrix $M \in \mathbb{R}^{N \times N}$ assigns $M_{ij} = \frac{1}{d_{S_i S_j}}$ for neighbors and 0 otherwise, giving higher weight to closer stations and reducing delay influence over longer distances~\cite{zhang2021train}. Further we hypothesize that the number of trains operating between a pair of stations significantly influences the extent of delay propagation. Specifically, when a large number of trains operate between two stations, opportunities for delay recovery become limited due to increased congestion and tighter scheduling constraints. In contrast, fewer trains between stations provide greater flexibility for compensating delays. To capture this intuition, we modify $M$ as 
\begin{equation}
M_{ij} = \frac{1}{d_{S_iS_j}} \times \frac{k_{S_iS_j}}{k_{max}}
\end{equation}
where $d_{S_i S_j}$ denotes the distance between stations $S_i$ and $S_j$, $k_{S_i S_j}$ is the number of trains operating between them, and $k_{max}$ is the maximum train count among all station pairs used for normalization. Finally, the spatial attention matrix is obtained as $Q = Softmax(C \odot M)$, where $\odot$ denotes element-wise multiplication.

\vspace{2mm}
\noindent \textit{Graph convolution:} The node encodings from the spatial attention module are fed into a graph convolutional network (GCN) to model structural dependencies among stations. Following \cite{zhang2021train}, graph signals are transformed into the spectral domain using Chebyshev polynomials, where graph convolution aggregates information from neighboring nodes to produce enriched node representations.


\noindent \textit{2D-CNN:} After the GCN network, a two-dimentional CNN is employed to merge the information of the nodes for adjacent time slices. For example, the convolution at the \( r \)-th layer of the daily component is given by:
\[
X_d^{(r)} = \mathrm{ReLU} \left( \Phi * \left( \mathrm{ReLU} \left( g_\theta *_{G} \hat{X}_d^{(r-1)} \right) \right) \right)
\]
where $\Phi$ is the convolution kernel parameter. 

\vspace{2mm}
\noindent \textit{Integration of all components:}
During fusion, each node assigns different importance to the three components, with weights learned from historical train data. The final integrated output is:
$\hat{\mathcal{Y}} = W_h \odot \hat{Y}_h + W_d \odot \hat{Y}_d + W_w \odot \hat{Y}_w $ 
where \( W_h \), \( W_d \), and \( W_w \) are learnable parameters. \(\hat{Y}_h\), \(\hat{Y}_d\), and \(\hat{Y}_w\) denote the final outputs of the recent, daily, and weekly components after the 2D-CNN.

\vspace{2mm}
\noindent \textit{Final Layer:}
To ensure non-negative outcomes to predict the average arrival delay, we propose to employ a ReLU operator in the final outcome of this layer. We modify $\hat{Y}$ as:
$\hat{Y} = ReLU(\hat{\mathcal{Y}})$.
In the Indian railway system, trains typically arrive either on time or with a delay, while early arrivals are extremely rare. To better reflect this operational reality and reduce noise in the data, we eliminate negative delay values at prediction time. This ensures that the model focuses solely on predicting delays, aligning with real-world patterns, and improving the robustness of the prediction task.


%% file: 040Experiment.tex
\section{Experimental Evaluation}

We validate our model by performing several experiments over the Indian Railway network (introduced in Section~\ref{sec:data}).

\subsection{Data processing and splits}


We preprocess the data by converting delays from minutes to hours and computing the hourly average arrival delay for each station over a period of $30 \times 24 = 720$ hours. Hours with no train arrivals at a station are excluded from prediction targets but retained in the historical records to maintain temporal consistency. Extreme delay cases ($\geq 24$ hours) are preserved to reflect the true distribution of operational disruptions.


\textit{Data splits:}For evaluation, the first $168$ timestamps (the initial $7$ days, up to $8$ September $2024, 12:00$ AM) are used exclusively as historical context and are excluded from training and testing. The remaining $23$ days of data are split chronologically into 70\% training, 10\% validation, and 20\% testing sets. The test set begins on $26$ September $2024$ at $08:00$ AM. We employ a sliding-window strategy to incorporate past delay history into the model input.


%

\subsection{Metrics}


To evaluate model performance, we use three standard metrics: (1)~Mean Absolute Error (MAE), (2)~Mean Absolute Percentage Error (MAPE), and (3)~Root Mean Square Error (RMSE). For MAPE computation, division-by-zero cases are handled separately: when the actual delay is zero, we compute the percentage difference directly between the predicted and actual values. All reported metrics are averaged across all stations in the network/zone and all timestamps in the test set.


\subsection{Baselines}

\ck{We compare RSTGCN against 9 baselines spanning statistical, machine learning, deep learning, and graph-based approaches. These include the statistical approach
(i)~Historical Average (HA); recurrent neural networks such as (ii)~LSTM~\cite{fu2016using, huang2020deep} and 
(iii)~GRU~\cite{gu2019improved, fu2016using}; 
traditional machine learning models including (iv)~Artificial Neural Networks (ANN)~\cite{wang2003artificial, yaghini2013railway}, 
(v)~Support Vector Regression (SVR)~\cite{awad2015support}, and (vi)~Random Forest (RF)~\cite{nabian2019predicting, nair2019ensemble}; 
and graph-based spatio-temporal models, namely 
(vii)~STGCN~\cite{heglund2020railway}, 
(viii)~ASTGCN~\cite{guo2019attention}, and 
(ix)~TSTGCN~\cite{zhang2021train}, which capture spatial correlations and temporal dynamics in railway networks, with attention mechanisms incorporated in the last two.}

\subsection{\ck{Experimental setup and hyper-parameters}} 

Table~\ref{tab:baselines} presents the range of hyper-parameters for the \textit{non-graph baselines} \ck{(LSTM, GRU, ANN, RF, and SVR)}. To ensure fairness, we select hyper-parameters for each zone distinctly, based on the best performance over validation data.

\begin{table}[htbp]
\centering
\scriptsize
\caption{\ck{Hyper-parameters for non-graph baselines}}
\resizebox{0.49\textwidth}{!}{
\begin{tabular}{l l l}
\hline
\textbf{LSTM} & \textbf{GRU} & \textbf{SVR} \\
\hline
Hidden Units: [50, 75, 100, 150] &
Units: [32, 64, 100] &
Regularization(C): [1, 10] \\

LR: [0.001, 0.0005, 0.0002] &
LR: [0.001, 0.0005] &
Epsilon: [0.01, 0.1] \\

Batch Size: [32, 64, 128] &
Batch Size: [64, 128] &
Kernel: [rbf, linear] \\

&
Dropout: [0, 0.2] &
Gamma: [scale, auto] \\
\midrule


\textbf{ANN} & \textbf{RF} & \\
\hline
Hidden$_1$: [128, 256] &
N Estimators: [100, 200, 400] & \\

Hidden$_2$: [64, 128] &
Max Depth: [10, 20, None] & \\

Dropout: [0, 0.2] &
Min Samples Split: [2, 5, 10] & \\

LR: [0.001, 0.0005] &
Min Samples Leaf: [1, 2, 4] & \\

Batch Size: [64, 128] &
Max Features: [sqrt, log2] & \\
\hline
\end{tabular}
}
\label{tab:baselines}
\end{table}

\ck{\textit{Graph-based models} (STGCN, ASTGCN, TSTGCN, and RSTGCN)
are implemented with the MXNet package (\url{https://mxnet.apache.org/}) on a system with $48$ CPU cores. To ensure fairness, we adopt an identical architectural configuration for all these models, which is adopted from TSTGCN~\cite{zhang2021train}.
The order of Chebyshev polynomials is set to $3$, the batch size is set as $4$, and all convolutional layers utilize $64$ kernels. Each temporal convolution layer also employs $64$ kernels.}



\ck{For temporal input segmentation, we set the lengths of the three components as $T_h = 3$, $T_d = 1$, and $T_w = 1$. 
The size of the prediction window is set to $T_p = 1, 2, 3$, which means that the model is trained to predict the aggregated arrival delays for each station for the next $1$, $2$ and $3$ hours, using data from the previous three hours, the corresponding hour of the previous day, and the same hour of the previous week.
Models are trained using mean squared error (MSE) as the loss function, optimized via backpropagation. During training, 
a learning rate of $0.001$ is used with number of epochs as 100. 
Early stopping is used to stop overfitting for all models.}

\subsection{Delay prediction results}


\ck{Table~\ref{tab:avg_model_zones} presents the average performance of all 10 models (9 baselines and our proposed model) across all zones for the three evaluation metrics, providing an overall comparison of their effectiveness. We see that our proposed RSTGCN achieves the best overall performance. 
}

\ck{Detailed zone-wise results for the seven strongest baselines alongside \our\ are reported in Tables~\ref{tab:zone_mae}, \ref{tab:zone_mape}, and \ref{tab:zone_rmse} for the three metrics, MAE, MAPE, and RMSE, respectively. 
In these tables, the zones are ordered by decreasing number of stations.
For readability, LSTM and ANN are excluded from the detailed zone-wise comparisons, as these two models generally under-perform the other baselines across most zones.}

\begin{table}[tb]
\centering
\caption{\ck{Average performance of all the models (averaged across all railway zones) for 1-hour / 2-hour / 3-hour prediction horizons. Best results are highlighted in \textbf{bold}.}}
\resizebox{\columnwidth}{!}{
\begin{tabular}{l|c|c|c}
\hline
\textbf{Model} & \textbf{MAE} & \textbf{MAPE (\%)} & \textbf{RMSE} \\
\hline
HA &
0.151 / 0.131 / 0.125 &
29.62 / 25.28 / 23.89 &
0.617 / 0.564 / 0.548 \\

LSTM &
0.160 / 0.150 / 0.145 &
35.68 / 32.60 / 30.80 &
0.671 / 0.630 / 0.613 \\

ANN &
0.139 / 0.122 / 0.117 &
26.61 / 23.17 / 21.88 &
0.597 / 0.538 / 0.520 \\

GRU &
0.138 / 0.121 / 0.117 &
26.17 / 23.13 / 22.30 &
0.589 / 0.529 / 0.512 \\

SVR &
0.124 / 0.109 / 0.103 &
19.79 / 17.17 / 16.20 &
0.563 / 0.505 / 0.488 \\

RF &
0.123 / 0.111 / 0.107 &
23.76 / 21.32 / 20.46 &
0.540 / 0.488 / 0.472 \\

STGCN &
0.098 / 0.099 / 0.099 &
16.66 / 17.21 / 17.11 &
0.477 / 0.471 / 0.472 \\

ASTGCN &
0.104 / 0.104 / 0.104 &
12.36 / 12.70 / 12.87 &
0.453 / 0.413 / 0.402 \\

TSTGCN &
0.090 / 0.097 / 0.099 &
10.09 / 11.51 / 11.95 &
0.452 / 0.411 / 0.399 \\

\textbf{RSTGCN} &
\textbf{0.074 / 0.081 / 0.085} &
\textbf{8.53 / 9.80 / 10.37} &
\textbf{0.427 / 0.393 / 0.387} \\
\hline
\end{tabular}}
\label{tab:avg_model_zones}
\end{table}

\input{Tables/zones}




\ck{Among the baselines, traditional temporal-only models (HA, LSTM, GRU, ANN, RF, and SVR) consistently perform worse than graph-based spatio-temporal models. HA exhibits the weakest overall performance. LSTM improves modestly over HA but is limited by its inability to capture spatial dependencies. GRU and ANN achieve comparable performance and outperform LSTM, yet both remain constrained by the lack of spatial modeling. RF further improves predictive accuracy, while SVR generally surpasses RF across most settings, demonstrating the effectiveness of nonlinear regression; however, both methods still fall short of the graph-based deep learning approaches that explicitly model spatio-temporal interactions.
}


{Graph-based models, namely STGCN, ASTGCN, and TSTGCN, significantly outperform the traditional baselines by leveraging spatio-temporal relationships. STGCN provides stable performance but is limited by the lack of an attention mechanism. ASTGCN improves over STGCN through attention-based long-term dependency modeling. TSTGCN further improves upon ASTGCN and achieves the best baseline performance by incorporating inter-station distance information, highlighting the importance of appropriate features.}


{As evident from the results, \textit{RSTGCN consistently achieves the lowest prediction errors across most zones and forecasting horizons}. Its strong performance across all 3 horizons demonstrates robust generalization across temporal settings and zones. This can be attributed to the incorporation of inter-station distances, train frequency, and the newly proposed features.}





{Table~\ref{tab:main} reports the forecasting performance of RSTGCN, TSTGCN, and ASTGCN for the next $1$, $2$, and $3$ hours on the full IRN.} 
{RSTGCN consistently outperforms the baselines for both individual horizons and aggregate performance, with \kc{18--21\% improvement in MAE, 14--24\% in MAPE, and 1--8\% in RMSE}. These results further demonstrate that the benefits of the proposed features and architectural modifications extend effectively to the large-scale IRN setting.}



\input{Tables/IRN}

\subsection{Long-horizon forecasting}

 \begin{figure*}[ht]
     \centering	
      \subfloat   {\includegraphics[width=0.3\linewidth]
     {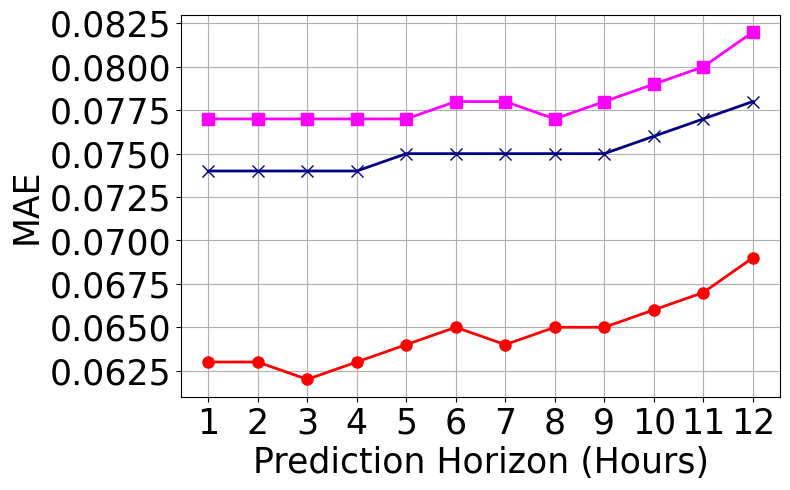}}
      \subfloat      {\includegraphics[width=0.3\linewidth]
      {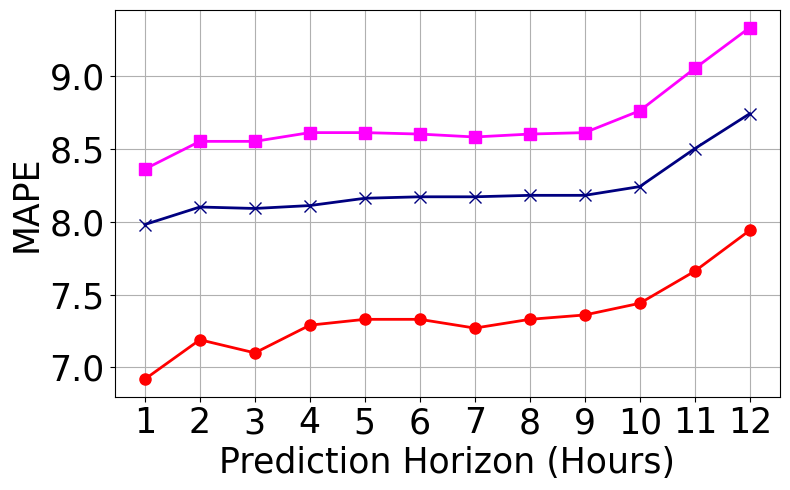}}
      \subfloat      {\includegraphics[width=0.4\linewidth, height=3.3cm]
      {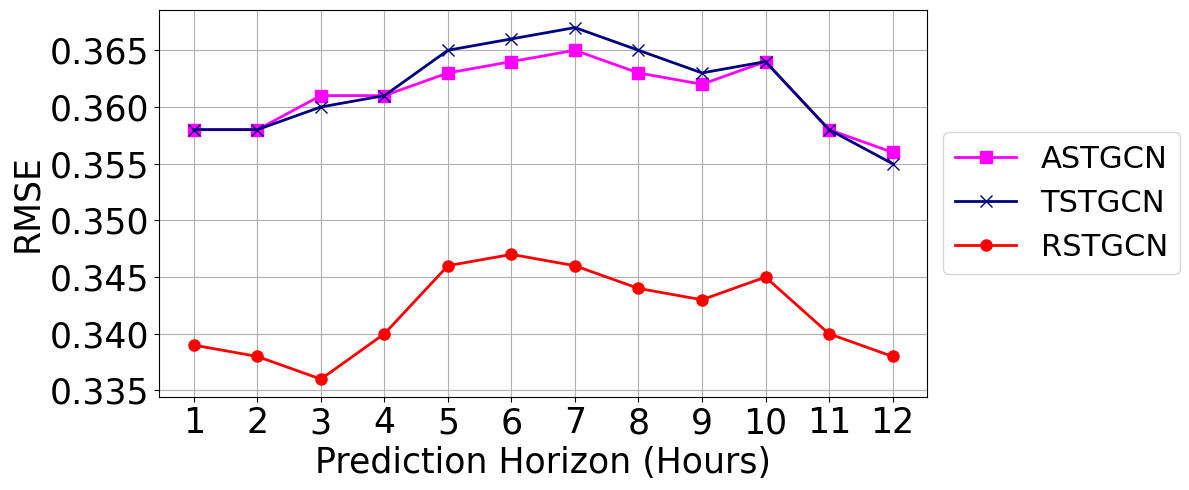}}
     \caption{Delay prediction performance of \our with two best baselines on the largest zone (SCR) for long-term prediction (up to $12$ hours) for all the metrics (averaged over all stations in zone SCR). \our consistently outperforms baselines.}
     \label{fig:longpred}
 \end{figure*}


{We further evaluate the long-horizon forecasting performance (1--12 hours) of RSTGCN and the next best baselines, ASTGCN and TSTGCN, on SCR, the largest zone with 499 stations. Figure~\ref{fig:longpred} reports MAE, MAPE, and RMSE as the prediction horizon varies from $1$ to $12$ hours, where the metrics for the $i$-th hour are averaged over horizons from $1$ to $i$.
We observe that the prediction error of all models increases with longer horizons. However, RSTGCN controls this degradation more effectively and consistently outperforms the baselines across all horizons.}

\subsection{\kc{Error analysis}}

\begin{table}[tb]
\centering
\scriptsize
\caption{\kc{Error analysis: Predictions of our model and  best baselines for 1-hour, 2-hour, 3-hour forecasting horizon, shown for some selected stations and timestamps. The predicted delays that are closest to the ground truth are in bold. The top row shows stable or increasing delays where \our performs well, whereas the bottom row shows irregular fluctuations where other models perform better at times.}}
\resizebox{.49\textwidth}{!}{
\begin{tabular}{c c c}
\hline
\textbf{Model} & \textbf{[1-hr, 2-hr, 3-hr] delays} & \textbf{[1-hr, 2-hr, 3-hr] delays} \\
\hline
& \textbf{\underline{NDLS -- NR Zone}} & \textbf{\underline{MDU -- SR Zone}} \\
\textbf{Ground Truth} & [0.22, 1.43, 1.87] & [0.32, 0.05, 0.10] \\
\textbf{ASTGCN} & [0.17, 1.18, 0.92] & [0.24, 0.21, 0.29] \\
\textbf{TSTGCN} & [0.57, 1.68, 0.89] & [0.27, 0.17, 0.19] \\
\textbf{RSTGCN} & [\textbf{0.19, 1.57, 1.62}] & [\textbf{0.32, 0.08, 0.12}] \\

\hline 

& \textbf{\underline{GHY -- NFR Zone}} & \textbf{\underline{LJN -- NER Zone}} \\
\textbf{Ground Truth} & [0.29, 0.0, 4.11] & [1.15, 2.58, 0.0] \\
\textbf{ASTGCN} & [0.42, \textbf{0.35}, \textbf{0.69}] & [1.09, \textbf{0.25}, 0.21] \\
\textbf{TSTGCN} & [0.35, 0.63, 0.47] & [1.05, 0.22, \textbf{0.16}] \\
\textbf{RSTGCN} & [\textbf{0.28}, 0.72, 0.02] & [\textbf{1.17}, 0.24, 0.26] \\
\hline
\end{tabular}
}
\label{tab:pred}
\end{table}

Till now, we have presented results aggregated over all stations in a zone and the full IRN. 
Now we check the predictions for some individual stations.
Table~\ref{tab:pred} shows the ground truth mean arrival delays and the 
predictions of \our and the two best baselines for 1-hour, 2-hour, 3-hour forecasting horizon, for some selected stations and timestamps.

\textit{Simpler cases:} The top row shows cases where delays increase consistently with the forecasting horizon or remain relatively stable. In these scenarios, \our consistently outperforms the baselines.

\textit{Complex cases:} {The bottom row contains scenarios with irregular delay patterns and abrupt fluctuations. In such cases, TSTGCN or ASTGCN can sometimes outperform \our, although all models perform poorly (\textit{i.e.} predicted delays are very different from ground truths).}

{Since most station--timestamp instances in IRN follow the simpler trend patterns (the highly fluctuating cases being much less frequent), \our achieves superior overall performance on the test set.}



\subsection{Ablation Study}


We perform an ablation study to analyze the contributions of (1)~the proposed features and (2)~the architectural modifications in \our on two medium-sized regions: Southern Railway and North Eastern Railway. 

\input{Tables/ablations}

\textit{Feature variants:} 
{Table~\ref{tab:ablations} presents 3-step prediction results for \ourad (2 features: average arrival and departure delay), \ouradh (3 features: average arrival and departure delays and headway), \ouradtd (4 features: average and total arrival and departure delay), and RSTGCN$_{\text{AvgDelay+Headway+TotDelay}}$ (all 5 features). These variants are compared with the best baseline, TSTGCN. All RSTGCN variants include the proposed architectural modifications.}


{We observe that incorporating average delay improves performance over TSTGCN, while adding hourly headway and total delay further improves performance over \ourad, demonstrating the contribution of each feature. Among the proposed features, total delay provides the largest gain. Finally, RSTGCN$_{\text{AvgDelay+Headway+TotDelay}}$ achieves the best overall performance by combining all proposed features.}


\textit{Model Variants:} {Table~\ref{tab:model_variants} presents the impact of individual architectural modifications. We begin with \ourbase, a version of \our that includes all proposed features but no architectural changes. \oursattn and \ourrelu respectively incorporate the modified spatial attention and ReLU activation in the last layer over \ourbase. We observe that \ourrelu provides a larger performance gain than \oursattn. Finally, combining both the updated spatial attention mechanism and ReLU in \our achieves the best performance, demonstrating the benefit of using both modifications together.}

\subsection{Variation of delay thresholds}

Finally, we consider a setup where we distinguish between minor deviations and operationally meaningful delays by considering thresholds on the delay.
Specifically, we consider only those delays that are greater than a specified threshold (e.g., greater than 0, 5 or 10 minutes). 

\ck{Table~\ref{tab:thresholds} compares the performance of \our against the strongest baselines across 1-hour, 2-hour, and 3-hour forecasting horizons under three delay thresholds (0, 5, and 10 minutes) over two representative zones, SWR and ECR. RSTGCN consistently achieves superior performance across most settings. Although ASTGCN and TSTGCN occasionally attain competitive results at specific horizons, RSTGCN demonstrates most stable and robust performance across different delay severity levels and prediction windows.}

\begin{table*}[htbp]
\centering
\scriptsize
\caption{ \ck{
Performance comparison of \our with the best baselines upto 1-hour / 2-hour / 3-hour horizon, where delays that are greater than a specific threshold only are considered. Results shown for three thresholds -- delays greater than [0, 5, 10] minutes. The best model results are shown in \textbf{bold} and the second-best are \underline{underlined}.}}
\resizebox{\textwidth}{!}{
\begin{tabular}{c | c c c | c c c | c c c}
\hline

\textbf{Zone} 
& ASTGCN & TSTGCN & RSTGCN 
& ASTGCN & TSTGCN & RSTGCN
& ASTGCN & TSTGCN & RSTGCN \\
\hline

& \multicolumn{3}{c|}{\textbf{Delay \textgreater 0 min}}
& \multicolumn{3}{c|}{\textbf{Delay \textgreater 5 min}}
& \multicolumn{3}{c}{\textbf{Delay \textgreater 10 min}} \\

\midrule

\multicolumn{10}{c}{\textbf{\underline{MAE}}} \\

SWR 
& \underline{0.074} / 0.088 / 0.088 
& \underline{0.074} / \underline{0.084} / \underline{0.083} 
& \textbf{0.062} / \textbf{0.075} / \textbf{0.077}

& 0.389 / 0.429 / 0.421 
& \underline{0.385} / \underline{0.419} / \underline{0.407} 
& \textbf{0.371} / \textbf{0.413} / \textbf{0.404}

& 0.439 / 0.499 / 0.491 
& \underline{0.438} / \underline{0.488} / \textbf{0.475} 
& \textbf{0.425} / \textbf{0.479} / \underline{0.480} \\

ECR 
& 0.126 / 0.160 / 0.162 
& \underline{0.107} / \underline{0.152} / \underline{0.157} 
& \textbf{0.103} / \textbf{0.147} / \textbf{0.149}

& 0.615 / 0.786 / 0.802 
& \textbf{0.535} / \underline{0.752} / \textbf{0.778} 
& \underline{0.548} / \textbf{0.749} / \underline{0.800}

& 0.683 / 0.858 / 0.880 
& \textbf{0.597} / \textbf{0.819} / \underline{0.858} 
& \underline{0.647} / \underline{0.829} / \textbf{0.855} \\

\midrule

\multicolumn{10}{c}{\textbf{\underline{MAPE (\%)}}} \\


SWR 
& 16.69 / 20.46 / 20.16 
& \underline{15.73} / \underline{18.17} / \underline{18.32} 
& \textbf{12.07} / \textbf{16.23} / \textbf{16.77}

& 74.41 / 83.82 / 84.06 
& \underline{69.67} / \underline{76.81} / \underline{77.92} 
& \textbf{63.53} / \textbf{76.47} / \textbf{77.69}

& 66.30 / 75.33 / 76.82 
& \underline{63.57} / \underline{70.94} / \underline{71.77} 
& \textbf{61.60} / \textbf{70.83} / \textbf{72.34} \\

ECR 
& 19.11 / 25.51 / \underline{25.72} 
& \underline{15.17} / \underline{25.10} / 25.74 
& \textbf{14.20} / \textbf{22.44} / \textbf{23.07}

& 79.19 / 100.51 / 100.38 
& \underline{64.42} / \underline{93.19} / \underline{95.85} 
& \textbf{64.34} / \textbf{91.81} / \textbf{94.87}

& 72.67 / 85.80 / 87.41 
& \underline{59.95} / \underline{79.50} / \underline{83.79} 
& \textbf{59.67} / \textbf{78.80} / \textbf{82.94} \\

\midrule

\multicolumn{10}{c}{\textbf{\underline{RMSE}}} \\


SWR 
& \underline{0.331} / 0.354 / 0.355 
& 0.335 / \underline{0.350} / \underline{0.348} 
& \textbf{0.327} / \textbf{0.348} / \textbf{0.345}

& \textbf{0.799} / 0.858 / 0.860 
& 0.815 / \underline{0.855} / \underline{0.850} 
& \underline{0.811} / \textbf{0.853} / \textbf{0.847}

& \underline{0.870} / \underline{0.949} / 0.954 
& 0.887 / \underline{0.950} / \underline{0.948} 
& \textbf{0.867} / \textbf{0.948} / \textbf{0.946} \\

ECR 
& \underline{0.527} / \underline{0.672} / 0.674 
& \textbf{0.521} / 0.672 / \underline{0.673} 
& 0.562 / \textbf{0.669} / \textbf{0.672}

& 1.235 / \underline{1.612} / \underline{1.630} 
& \underline{1.233} / \textbf{1.611} / \textbf{1.627} 
& \textbf{1.230} / 1.623 / 1.643

& \underline{1.317} / \textbf{1.708} / 1.732 
& \textbf{1.312} / \underline{1.709} / \underline{1.730} 
& 1.390 / 1.735 / \textbf{1.728} \\

\hline
\end{tabular}
}
\label{tab:thresholds}
\end{table*}

%% file: Tables/zones.tex
\begin{table*}[!t]
\centering
\scriptsize
\caption{Zone-wise Mean Absolute Error (MAE) across different delay prediction models. Each row corresponds to a zone, and gives the aggregated average arrival delay error (in hours) over all stations within that zone. 
Each column is for a prediction model.
For each model, results are shown for three prediction horizons: upto 1-hour / 2-hour / 3-hour. Best values are in bold.}
\resizebox{\textwidth}{!}{
\begin{tabular}{l|c|c|c|c|c|c|c|c}
\hline
\textbf{Zones} & \textbf{HA} & \textbf{GRU} & \textbf{RF} & \textbf{SVR} & \textbf{STGCN} & \textbf{ASTGCN} & \textbf{TSTGCN} & \textbf{RSTGCN (ours)} \\
\hline
SCR  & 0.127 / 0.110 / 0.105 & 0.114 / 0.100 / 0.096 & 0.104 / 0.096 / 0.092 & 0.102 / 0.089 / 0.084 & 0.096 / 0.099 / 0.099 & 0.087 / 0.094 / 0.095 & 0.079 / 0.088 / 0.089 & \textbf{0.061 / 0.071 / 0.076} \\
NR   & 0.187 / 0.167 / 0.159 & 0.173 / 0.156 / 0.150 & 0.158 / 0.145 / 0.140 & 0.163 / 0.149 / 0.143 & 0.124 / 0.126 / 0.126 & 0.127 / 0.130 / 0.131 & 0.110 / 0.122 / 0.125 & \textbf{0.098 / 0.103 / 0.104} \\
WR   & 0.066 / 0.057 / 0.054 & 0.076 / 0.068 / 0.065 & 0.055 / 0.049 / 0.047 & 0.051 / 0.044 / 0.041 & 0.041 / 0.040 / 0.040 & 0.043 / 0.041 / 0.039 & 0.036 / 0.040 / 0.039 & \textbf{0.029 / 0.032 / 0.033} \\
ECR  & 0.188 / 0.166 / 0.158 & 0.197 / 0.173 / 0.167 & 0.155 / 0.142 / 0.137 & 0.161 / 0.144 / 0.139 & 0.114 / 0.113 / 0.114 & 0.120 / 0.128 / 0.128 & 0.106 / 0.123 / 0.126 & \textbf{0.095 / 0.106 / 0.108} \\
NWR  & 0.079 / 0.065 / 0.062 & 0.070 / 0.058 / 0.056 & 0.061 / 0.053 / 0.050 & 0.063 / 0.051 / 0.047 & 0.041 / 0.041 / 0.041 & 0.043 / 0.041 / 0.039 & 0.041 / 0.041 / 0.040 & \textbf{0.035 / 0.035 / 0.034} \\
SR   & 0.117 / 0.107 / 0.104 & 0.106 / 0.099 / 0.097 & 0.092 / 0.088 / 0.087 & 0.085 / 0.078 / 0.076 & 0.090 / 0.094 / 0.094 & 0.078 / 0.084 / 0.087 & 0.068 / 0.083 / 0.088 & \textbf{0.054 / 0.063 / 0.067} \\
NFR  & 0.098 / 0.081 / 0.075 & 0.087 / 0.072 / 0.067 & 0.082 / 0.071 / 0.067 & 0.084 / 0.073 / 0.069 & 0.052 / 0.052 / 0.052 & 0.062 / 0.057 / 0.055 & 0.054 / 0.054 / 0.054 & \textbf{0.042 / 0.046 / 0.046} \\
CR   & 0.193 / 0.172 / 0.164 & 0.170 / 0.158 / 0.156 & 0.165 / 0.153 / 0.149 & 0.156 / 0.139 / 0.131 & 0.140 / 0.145 / 0.146 & 0.142 / 0.146 / 0.147 & 0.109 / 0.131 / 0.137 & \textbf{0.102 / 0.110 / 0.116} \\
ECOR & 0.131 / 0.110 / 0.104 & 0.122 / 0.104 / 0.099 & 0.108 / 0.097 / 0.092 & 0.107 / 0.090 / 0.084 & 0.084 / 0.085 / 0.084 & 0.092 / 0.091 / 0.091 & 0.080 / 0.084 / 0.087 & \textbf{0.065 / 0.072 / 0.073} \\
NCR  & 0.185 / 0.157 / 0.149 & 0.162 / 0.139 / 0.133 & 0.148 / 0.133 / 0.128 & 0.147 / 0.127 / 0.119 & 0.115 / 0.116 / 0.116 & 0.121 / 0.120 / 0.125 & 0.110 / 0.113 / 0.115 & \textbf{0.096 / 0.104 / 0.109} \\
SWR  & 0.077 / 0.066 / 0.063 & 0.068 / 0.059 / 0.057 & 0.060 / 0.054 / 0.051 & 0.065 / 0.059 / 0.057 & 0.050 / 0.049 / 0.049 & 0.049 / 0.052 / 0.052 & 0.047 / 0.048 / 0.050 & \textbf{0.034 / 0.036 / 0.039} \\
ER   & 0.109 / 0.097 / 0.093 & 0.109 / 0.100 / 0.098 & 0.087 / 0.079 / 0.076 & 0.091 / 0.082 / 0.079 & 0.056 / 0.057 / 0.057 & 0.071 / 0.070 / 0.069 & 0.057 / 0.062 / 0.063 & \textbf{0.042 / 0.048 / 0.048} \\
NER  & 0.189 / 0.165 / 0.158 & 0.163 / 0.141 / 0.135 & 0.152 / 0.133 / 0.127 & 0.154 / 0.132 / 0.126 & 0.095 / 0.094 / 0.094 & 0.108 / 0.102 / 0.103 & 0.099 / 0.098 / 0.100 & \textbf{0.078 / 0.079 / 0.084} \\
WCR  & 0.189 / 0.167 / 0.161 & 0.168 / 0.150 / 0.146 & 0.156 / 0.143 / 0.139 & 0.153 / 0.136 / 0.130 & 0.152 / 0.151 / 0.150 & 0.146 / 0.147 / 0.145 & 0.116 / 0.132 / 0.136 & \textbf{0.098 / 0.114 / 0.118} \\
SER  & 0.172 / 0.147 / 0.138 & 0.158 / 0.136 / 0.129 & 0.147 / 0.128 / 0.121 & 0.152 / 0.131 / 0.124 & 0.095 / 0.097 / 0.097 & 0.116 / 0.111 / 0.111 & 0.106 / 0.106 / 0.108 & \textbf{0.088 / 0.090 / 0.095} \\
SECR & 0.238 / 0.207 / 0.199 & 0.214 / 0.189 / 0.183 & 0.197 / 0.179 / 0.175 & 0.200 / 0.167 / 0.158 & 0.169 / 0.167 / 0.167 & 0.176 / 0.165 / 0.169 & 0.138 / 0.135 / 0.143 & \textbf{0.113 / 0.120 / 0.134} \\
KRCL & 0.217 / 0.185 / 0.174 & 0.184 / 0.160 / 0.153 & 0.165 / 0.151 / 0.146 & 0.177 / 0.159 / \textbf{0.144} & 0.160 / 0.162 / 0.159 & 0.182 / 0.186 / 0.189 & 0.169 / 0.184 / 0.185 & \textbf{0.133 / 0.156} / 0.163 \\
\hline
\end{tabular}}
\label{tab:zone_mae}
\end{table*}

\begin{table*}[!t]
\centering
\scriptsize
\caption{Zone-wise Mean Absolute Percentage Error (MAPE \%) across different delay prediction models}
\resizebox{\textwidth}{!}{
\begin{tabular}{l|c|c|c|c|c|c|c|c}
\hline
\textbf{Zones} & \textbf{HA} & \textbf{GRU} & \textbf{RF} & \textbf{SVR} & \textbf{STGCN} & \textbf{ASTGCN} & \textbf{TSTGCN} & \textbf{RSTGCN (ours)} \\
\hline
SCR  & 24.09 / 20.63 / 19.37 & 20.22 / 17.80 / 17.00 & 20.63 / 18.52 / 17.72 & 15.45 / 13.35 / 12.49 & 15.29 / 16.29 / 16.24 & 10.15 / 11.22 / 11.56 & 8.83 / 10.30 / 10.59 & \textbf{7.11 / 8.51 / 9.41} \\
NR   & 41.82 / 36.17 / 34.02 & 40.16 / 35.05 / 33.54 & 36.29 / 32.58 / 31.16 & 37.15 / 33.07 / 31.57 & 25.01 / 26.29 / 26.12 & 17.16 / 17.89 / 18.14 & 13.39 / 15.89 / 16.70 & \textbf{12.22 / 13.48 / 14.05} \\
WR   & 20.17 / 16.84 / 15.85 & 26.20 / 22.78 / 21.43 & 16.55 / 14.03 / 13.26 & 11.82 / 9.53 / 8.75 & 8.51 / 8.78 / 8.64 & 6.59 / 6.31 / 6.11 & 4.87 / 5.80 / 5.88 & \textbf{4.67 / 5.13 / 5.41} \\
ECR  & 35.04 / 30.52 / 29.26 & 43.07 / 36.95 / 34.84 & 29.85 / 27.10 / 26.15 & 30.25 / 27.09 / 25.91 & 19.95 / 20.78 / 20.68 & 13.32 / 15.10 / 15.15 & 11.54 / 14.71 / 15.04 & \textbf{10.19 / 12.50 / 12.73} \\
NWR  & 20.51 / 16.06 / 14.88 & 16.50 / 13.30 / 12.40 & 13.83 / 11.42 / 10.64 & 12.29 / 9.62 / 8.66 & 8.75 / 9.11 / 8.99 & 6.65 / 6.13 / 5.82 & 5.63 / 5.70 / 5.70 & \textbf{5.47 / 5.33 / 5.22} \\
SR   & 35.96 / 31.79 / 30.34 & 32.25 / 28.93 / 28.14 & 30.16 / 27.69 / 26.93 & 20.55 / 18.08 / 17.32 & 22.22 / 23.74 / 23.86 & 12.43 / 14.29 / 14.59 & 9.57 / 13.72 / 14.50 & \textbf{8.74 / 11.07 / 11.52} \\
NFR  & 17.96 / 14.31 / 13.01 & 12.41 / 10.43 / 10.01 & 14.35 / 12.02 / 11.15 & 14.11 / 11.85 / 10.97 & 7.01 / 7.03 / 7.22 & 7.07 / 6.57 / 6.27 & 5.94 / 6.04 / 6.14 & \textbf{4.72 / 5.35 / 5.29} \\
CR   & 37.06 / 32.80 / 31.24 & 26.98 / 27.14 / 28.52 & 33.42 / 31.06 / 30.38 & 22.08 / 19.79 / 18.94 & 26.24 / 27.92 / 27.54 & 17.03 / 17.74 / 17.87 & 12.09 / 15.42 / 16.31 & \textbf{11.46 / 12.99 / 13.81} \\
ECOR & 25.74 / 21.72 / 20.57 & 24.48 / 21.53 / 20.68 & 22.54 / 20.43 / 19.61 & 16.11 / 13.66 / 12.72 & 15.52 / 15.45 / 15.15 & 11.06 / 11.08 / 11.17 & 9.10 / 9.79 / 10.49 & \textbf{7.47 / 8.59 / 8.94} \\
NCR  & 31.08 / 26.33 / 25.07 & 25.32 / 22.43 / 21.62 & 24.77 / 22.46 / 21.56 & 18.87 / 16.63 / 15.67 & 17.43 / 17.64 / 17.40 & 13.12 / 13.03 / 14.04 & 11.48 / 11.86 / 12.52 & \textbf{9.97 / 10.96 / 11.93} \\
SWR  & 20.23 / 16.52 / 15.09 & 14.72 / 12.63 / 12.01 & 13.67 / 11.77 / 11.00 & 15.99 / 13.63 / 12.70 & 8.36 / 8.85 / 8.78 & 6.95 / 7.72 / 7.85 & 6.48 / 6.94 / 7.23 & \textbf{4.69 / 5.47 / 5.95} \\
ER   & 31.42 / 26.59 / 24.73 & 31.01 / 27.36 / 26.47 & 20.66 / 18.34 / 17.66 & 22.15 / 19.15 / 18.22 & 11.08 / 12.34 / 12.07 & 9.61 / 9.76 / 9.88 & 7.01 / 8.33 / 8.68 & \textbf{5.86 / 6.93 / 7.18} \\
NER  & 31.89 / 26.49 / 25.05 & 21.98 / 19.12 / 18.17 & 19.92 / 17.34 / 16.32 & 16.93 / 14.22 / 13.16 & 14.73 / 14.53 / 14.85 & 11.30 / 11.20 / 11.02 & 10.04 / 10.57 / 10.57 & \textbf{7.97 / 8.59 / 8.89} \\
WCR  & 36.45 / 32.08 / 30.75 & 30.61 / 27.46 / 26.46 & 30.72 / 27.85 / 27.02 & 22.00 / 19.33 / 18.36 & 24.26 / 25.04 / 25.33 & 17.45 / 17.56 / 17.78 & 13.30 / 15.64 / 16.62 & \textbf{10.98 / 13.25 / 14.50} \\
SER  & 25.41 / 21.66 / 20.79 & 20.99 / 18.12 / 17.48 & 19.80 / 17.48 / 16.72 & 20.12 / 17.64 / 16.78 & 14.35 / 14.84 / 14.74 & 12.04 / 11.55 / 11.59 & 10.82 / 11.03 / 11.18 & \textbf{8.88 / 9.19 / 9.64} \\
SECR & 44.00 / 38.28 / 36.23 & 37.74 / 34.60 / 33.34 & 38.49 / 35.51 / 34.17 & 23.04 / 20.03 / 18.88 & 28.96 / 28.05 / 27.52 & 19.24 / 19.51 / 20.64 & 14.05 / 15.17 / 16.35 & \textbf{10.92 / 13.32 / 15.33} \\
KRCL & 24.74 / 20.92 / 19.81 & 20.19 / 17.61 / 16.96 & 18.30 / 16.84 / 16.44 & 17.53 / \textbf{15.30 / 14.36} & 15.51 / 15.86 / 15.76 & 18.90 / 19.17 / 19.31 & 17.41 / 18.72 / 18.72 & \textbf{13.61} / 15.95 / 16.57 \\
\hline
\end{tabular}}
\label{tab:zone_mape}
\end{table*}

\begin{table*}[!t]
\centering
\scriptsize
\caption{Zone-wise Root Mean Square Error (RMSE) (in hours) across different delay prediction models}
\resizebox{\textwidth}{!}{
\begin{tabular}{l|c|c|c|c|c|c|c|c}
\hline
\textbf{Zones} & \textbf{HA} & \textbf{GRU} & \textbf{RF} & \textbf{SVR} & \textbf{STGCN} & \textbf{ASTGCN} & \textbf{TSTGCN} & \textbf{RSTGCN (ours)} \\
\hline
SCR  & 0.445 / 0.417 / 0.408 & 0.414 / 0.378 / 0.367 & 0.389 / 0.363 / 0.354 & 0.408 / 0.376 / 0.365 & 0.456 / 0.449 / 0.450 & 0.364 / 0.339 / 0.333 & 0.362 / 0.337 / 0.328 & \textbf{0.328 / 0.309 / 0.311} \\
NR   & 0.714 / 0.666 / 0.645 & 0.665 / 0.612 / 0.590 & 0.634 / 0.585 / 0.563 & 0.643 / 0.593 / 0.572 & 0.493 / 0.511 / 0.518 & 0.536 / 0.485 / 0.461 & 0.538 / 0.488 / 0.463 & \textbf{0.505 / 0.463 / 0.441} \\
WR   & 0.361 / 0.332 / 0.325 & 0.525 / 0.488 / 0.475 & 0.314 / 0.286 / 0.277 & 0.329 / 0.297 / 0.285 & 0.315 / 0.293 / 0.296 & 0.232 / 0.223 / 0.209 & 0.242 / 0.232 / 0.216 & \textbf{0.219 / 0.215 / 0.205} \\
ECR  & 0.827 / 0.751 / 0.730 & 0.840 / 0.750 / 0.724 & 0.712 / 0.642 / 0.623 & 0.736 / 0.661 / 0.641 & \textbf{0.524 / 0.515 / 0.530} & 0.548 / 0.521 / 0.494 & 0.552 / 0.528 / 0.503 & 0.548 / 0.513 / 0.486 \\
NWR  & 0.407 / 0.354 / 0.346 & 0.378 / 0.318 / 0.308 & 0.364 / 0.307 / 0.299 & 0.376 / 0.316 / 0.305 & 0.256 / 0.253 / 0.253 & \textbf{0.235 / 0.230 / 0.226} & 0.266 / 0.250 / 0.242 & 0.245 / 0.235 / 0.230 \\
SR   & 0.547 / 0.521 / 0.515 & 0.503 / 0.474 / 0.467 & 0.453 / 0.431 / 0.427 & 0.481 / 0.454 / 0.447 & 0.478 / 0.494 / 0.484 & 0.337 / 0.308 / 0.321 & 0.376 / 0.343 / 0.352 & \textbf{0.332 / 0.297 / 0.310} \\
NFR  & 0.460 / 0.411 / 0.393 & 0.434 / 0.377 / 0.354 & 0.411 / 0.363 / 0.344 & 0.415 / 0.371 / 0.353 & 0.477 / 0.414 / 0.386 & 0.339 / 0.300 / \textbf{0.275} & 0.358 / 0.310 / 0.285 & \textbf{0.319 / 0.302} / 0.282 \\
CR   & 0.717 / 0.668 / 0.644 & 0.674 / 0.618 / 0.598 & 0.636 / 0.587 / 0.568 & 0.660 / 0.601 / 0.574 & 0.640 / 0.661 / 0.678 & 0.556 / 0.517 / 0.506 & 0.512 / 0.490 / 0.487 & \textbf{0.510 / 0.484 / 0.478} \\
ECOR & 0.468 / 0.428 / 0.416 & 0.432 / 0.390 / 0.378 & 0.402 / 0.368 / 0.359 & 0.428 / 0.385 / 0.372 & 0.412 / 0.417 / 0.412 & 0.401 / 0.359 / 0.342 & 0.379 / 0.343 / 0.330 & \textbf{0.366 / 0.335 / 0.318} \\
NCR  & 0.773 / 0.692 / 0.663 & 0.705 / 0.618 / 0.585 & 0.675 / 0.597 / 0.568 & 0.709 / 0.621 / 0.588 & 0.522 / 0.521 / 0.527 & 0.538 / 0.469 / 0.475 & 0.547 / 0.476 / 0.472 & \textbf{0.530 / 0.464 / 0.469} \\
SWR  & 0.355 / 0.318 / 0.307 & 0.331 / 0.290 / 0.277 & 0.310 / 0.274 / 0.264 & 0.311 / 0.280 / 0.272 & 0.320 / 0.285 / 0.272 & 0.249 / 0.234 / 0.229 & 0.256 / 0.231 / 0.223 & \textbf{0.237 / 0.220 / 0.216} \\
ER   & 0.654 / 0.616 / 0.600 & 0.632 / 0.597 / 0.588 & 0.561 / 0.522 / 0.509 & 0.563 / 0.526 / 0.515 & 0.430 / 0.389 / 0.406 & 0.453 / 0.407 / 0.393 & 0.437 / 0.389 / 0.376 & \textbf{0.427 / 0.385 / 0.374} \\
NER  & 0.919 / 0.828 / 0.822 & 0.870 / 0.760 / 0.743 & 0.853 / 0.746 / 0.732 & 0.878 / 0.768 / 0.752 & \textbf{0.500 / 0.474 / 0.469} & 0.596 / 0.523 / 0.515 & 0.611 / 0.533 / 0.525 & 0.559 / 0.497 / 0.497 \\
WCR  & 0.666 / 0.617 / 0.606 & 0.614 / 0.560 / 0.546 & 0.581 / 0.536 / 0.524 & 0.614 / 0.560 / 0.545 & 0.609 / 0.613 / 0.611 & 0.528 / 0.497 / 0.478 & 0.518 / 0.489 / 0.472 & \textbf{0.501 / 0.476 / 0.463} \\
SER  & 0.670 / 0.605 / 0.581 & 0.628 / 0.555 / 0.531 & 0.597 / 0.528 / 0.504 & 0.612 / 0.544 / 0.522 & \textbf{0.433 / 0.449} / 0.460 & 0.544 / 0.478 / 0.466 & 0.530 / 0.466 / \textbf{0.454} & 0.502 / 0.448 / 0.449 \\
SECR & 0.831 / 0.757 / 0.733 & 0.766 / 0.685 / 0.660 & 0.713 / 0.647 / 0.628 & 0.791 / 0.700 / 0.673 & 0.698 / 0.688 / 0.695 & 0.664 / 0.581 / 0.577 & 0.629 / 0.543 / 0.544 & \textbf{0.594 / 0.531 / 0.541} \\
KRCL & 0.680 / 0.600 / 0.575 & 0.606 / 0.527 / 0.505 & 0.577 / 0.507 / 0.487 & 0.617 / 0.533 / 0.507 & 0.561 / 0.588 / 0.570 & 0.589 / 0.545 / 0.528 & 0.568 / 0.536 / 0.512 & \textbf{0.532 / 0.515 / 0.505} \\
\hline
\end{tabular}}
\label{tab:zone_rmse}
\end{table*}

%% file: Tables/IRN.tex
\begin{table}[htbp]
    \centering
    \scriptsize
    \caption{\kc{Forecasting performance over the entire IR Network for the next $1$, $2$, and $3$ hours with the best-performing $2$ baselines. Results are averaged over $3$ prediction steps, and (+ Impvt.) denotes the improvement achieved by RSTGCN over each baseline for the corresponding metric.}}
    \resizebox{0.5\textwidth}{!}{
    \begin{tabular}{l c c c}
    \toprule
    & \textbf{MAE} & \textbf{MAPE} & \textbf{RMSE} \\
    \midrule
    \textbf{ASTGCN} & 0.087 / 0.093 / 0.095 & 9.71 / 11.26 / 11.62 & 0.444 / 0.408 / 0.396 \\
    Avg + Impvt.  & 0.092 (+ \textbf{18.47\%}) & 10.86 (+ \textbf{14.91\%}) & 0.416 (+ \textbf{8.65\%}) \vspace{2pt} \\
    \textbf{TSTGCN} & 0.094 / 0.096 / 0.097 & 11.74 / 12.44 / 12.42 & 0.393 / 0.382 / 0.379 \\
    Avg + Impvt. & 0.096 (+ \textbf{21.86\%}) & 12.2 (+ \textbf{24.26\%}) & 0.385 (+ \textbf{1.29\%}) \vspace{2pt} \\
    \textbf{RSTGCN} & 0.069 / 0.078 / 0.079 & 8.12 / 9.71 / 9.89 & 0.389 / 0.378 / 0.373 \\
    Avg & \textbf{0.075} & \textbf{9.24} & \textbf{0.38} \\
    \bottomrule
    \end{tabular}
    }
    \label{tab:main}
\end{table}

%% file: Tables/ablations.tex
\begin{table*}[htbp]
\centering
\scriptsize
\caption{Ablation study with feature variants upto 1-hour / 2-hour / 3-hour forecasts in comparison with the best baseline. The subscripts in the column headers denote the features used in the corresponding variant of RSTGCN. All the variants of RSTGCN use the updated spatial attention and ReLU function. The best results are shown in \textbf{bold}, the second-best are \underline{underlined}, and the third-best are \textit{italicized}.}
\label{tab:ablations}
\resizebox{\textwidth}{!}{
\begin{tabular}{c | c | c | c | c | c}
\hline
\textbf{Zones} & \textbf{TSTGCN} & \textbf{\ourad} & \textbf{\ouradh} & \textbf{\ouradtd} & \textbf{RSTGCN$_{\text{AvgDelay+Headway+TotDelay}}$} \\
\hline
\multicolumn{6}{c}{\textbf{\underline{MAE}}} \\
SR  & 0.068 / 0.083 / 0.088 & 0.061 / 0.072 / 0.075 & \textit{0.059 / 0.067 / 0.070} & \underline{0.055 / 0.066 / 0.068} & \textbf{0.054 / 0.063 / 0.067} \\
NER & 0.099 / 0.098 / 0.100 & \textit{0.083 / 0.088} / 0.093 & \underline{0.081 / 0.082} \textit{/ 0.090} & \underline{0.081 / 0.082 / 0.086} & \textbf{0.078 / 0.079 / 0.084} \\
\multicolumn{6}{c}{\textbf{\underline{MAPE (\%)}}} \\
SR  & 9.57 / 13.72 / 14.50 & 9.47 / 12.67 / 12.85 & \textit{9.04} \underline{/ 11.12 / 11.66} & \textbf{8.58} \textit{/ 11.64 / 11.76} & \underline{8.74} \textbf{/ 11.07 / 11.52} \\
NER & 10.04 / 10.57 / 10.57 & 8.64 / 9.63 / 10.02 & \underline{8.29} \textit{/ 9.01 / 9.59} & \textit{8.36} \underline{/ 8.95 / 9.07} & \textbf{7.97 / 8.59 / 8.89} \\
\multicolumn{6}{c}{\textbf{\underline{RMSE}}} \\
SR  & 0.376 / 0.343 / 0.352 & 0.378 / 0.338 / 0.340 & \textit{0.361 / 0.318 / 0.325} & \textbf{0.329} \underline{/ 0.299 / 0.313} & \underline{0.332} \textbf{/ 0.297 / 0.310} \\
NER & 0.611 / 0.533 / 0.525 & 0.606 / 0.532 / 0.526 & \textit{0.581 / 0.512 / 0.510} & \underline{0.575 / 0.508 / 0.505} & \textbf{0.559 / 0.497 / 0.497} \\[2pt]
\hline
\end{tabular}
}
\end{table*}

\begin{table*}[htbp]
\centering
\scriptsize
\caption{Ablation study with model variants upto 1-hour / 2-hour / 3-hour forecasts in comparison with the best baseline. The subscripts in the column headers indicate the specific architectural modifications applied in the corresponding variant of RSTGCN. All the variants of RSTGCN use all five features.}
\label{tab:model_variants}
\resizebox{\textwidth}{!}{
\begin{tabular}{c | c | c | c | c | c}
\hline
\textbf{Zones} & \textbf{TSTGCN} & \textbf{RSTGCN$_{\text{Base}}$} & \textbf{RSTGCN$_{\text{SAttn}}$} & \textbf{RSTGCN$_{\text{ReLU}}$} & \textbf{RSTGCN$_{\text{SAtt+ReLU}}$} \\ 
\hline
\multicolumn{6}{c}{\textbf{\underline{MAE}}} \\
SR  & 0.068 / 0.083 / 0.088 & 0.071 / 0.078 / 0.081 & \textit{0.065 / 0.076 / 0.079} & \underline{0.060 / 0.070 / 0.072} & \textbf{0.054 / 0.063 / 0.067} \\
NER & 0.099 / 0.098 / 0.100 & 0.102 / 0.097 / 0.098 & \textit{0.098 / 0.096 / 0.098} & \underline{0.079 / 0.083 / 0.085} & \textbf{0.078 / 0.079 / 0.084} \\
\multicolumn{6}{c}{\textbf{\underline{MAPE (\%)}}} \\
SR  & 9.57 / 13.72 / 14.50 & 10.52 / 12.87 / 13.24 & \underline{9.12} \textit{/ 12.24 / 12.64} & \textit{9.69} \underline{/ 12.01 / 12.24} & \textbf{8.74 / 11.07 / 11.52} \\
NER & 10.10 / 10.57 / 10.57 & 10.32 / 10.45 / 10.36 & \textit{10.06 / 10.27 / 10.31} & \underline{8.06 / 9.14 / 8.95} & \textbf{7.97 / 8.59 / 8.89} \\
\multicolumn{6}{c}{\textbf{\underline{RMSE}}} \\
SR  & 0.374 / 0.343 / 0.352 & \textit{0.345 / 0.308} / 0.321 & 0.351 / 0.312 / \textit{0.320} & \underline{0.335 / 0.307 / 0.319} & \textbf{0.332 / 0.297 / 0.310} \\
NER & 0.611 / 0.533 / 0.525 & 0.608 / 0.529 / 0.519 & \textit{0.605 / 0.527 / 0.518} & \underline{0.565 / 0.506 / 0.503} & \textbf{0.559 / 0.497 / 0.497} \\ [2pt]
\hline
\end{tabular}
}
\end{table*}

%% file: 050Conclusion.tex
\section{Conclusion}
\label{sec:conclusion}

In this work, we address the problem of station-wise mean arrival delay prediction over large railway networks, with a focus on the Indian Railway system. We proposed \our, a spatio-temporal graph convolutional network that integrates domain-informed features into spatial dependency models to better capture delay propagation dynamics. 
Experiments on an Indian Railway delay dataset demonstrate that \our consistently outperforms strong baselines across multiple evaluation settings, achieving improvements of up to 18\% in MAE, 14\% in MAPE, and 1–8\% in RMSE on the IRN. 
Furthermore, the nationwide railway delay dataset introduced in this work will provide a strong foundation for future research.

\kc{\textit{Limitations and future work:} Our dataset spans one month only; including exogenous factors such as seasonal variations and festive travel patterns are potential future works.
Also, our dataset includes express trains only, and not freight and `local' (short distance) trains.
We observe a limited set of irregular and highly unpredictable delay scenarios in which performance degrades for all models, typically caused by sudden disruptions such as system failures;  addressing such extreme/unforeseen cases remains an important direction for future work. 
}